\documentclass[sigconf]{acmart}
\usepackage{makecell}

\DeclareMathOperator*{\argmin}{argmin}

\setcopyright{acmlicensed}
\acmISBN{} 
\acmDOI{} 

\acmConference[LBRS@RecSys '18]{RecSys 2018}{October 2-7, 2018}{Vancouver, BC, Canada}
\acmYear{2018}
\copyrightyear{2018}

\acmBooktitle{Proceedings of the Late-Breaking Results track part of the Twelfth ACM Conference on Recommender Systems}
\begin{document}
\title{A novel graph-based model for hybrid recommendations in cold-start scenarios}
%\titlenote{}
%\subtitle{Extended Abstract}
%\subtitlenote{}

\author{Cesare Bernardis}
%\authornote{}
%\orcid{1234-5678-9012}
\affiliation{%
  \institution{Politecnico di Milano, Italy}
  \streetaddress{Piazza Leonardo da Vinci, 32}
  %\city{Milano}
  %\state{Italy}
  \postcode{20133}
  \orcid{0000-0002-8972-0850}
}
\email{cesare.bernardis@mail.polimi.it}

\author{Maurizio Ferrari Dacrema}
%\authornote{}
\affiliation{%
  \institution{Politecnico di Milano, Italy}
  \streetaddress{Piazza Leonardo da Vinci, 32}
  %\city{Milano}
  %\state{Italy}
  \postcode{20133}
  \orcid{0000-0001-7103-2788}
}
\email{maurizio.ferrari@polimi.it}

\author{Paolo Cremonesi}
%\authornote{}
\affiliation{%
  \institution{Politecnico di Milano, Italy}
  \streetaddress{Piazza Leonardo da Vinci, 32}
  %\city{Milano}
  %\state{Italy}
  \postcode{20133}
  \orcid{0000-0002-1253-8081}
}
\email{paolo.cremonesi@polimi.it}

\begin{abstract}
Cold-start is a very common and still open problem in the Recommender Systems literature.
Since cold start items do not have any interaction, collaborative algorithms are not applicable. One of the main strategies is to use pure or hybrid content-based approaches, which usually yield to lower recommendation quality than collaborative ones.
Some techniques to optimize performance of this type of approaches have been studied in recent past.
One of them is called feature weighting, which assigns to every feature a real value, called weight, 
that estimates its importance.
Statistical techniques for feature weighting commonly used in Information Retrieval, like TF-IDF, 
have been adapted for Recommender Systems, but they often do not provide sufficient quality improvements.
More recent approaches\cite{fbsm2015, lfw} estimate weights by leveraging collaborative information via machine learning, in order to learn the importance of a feature based on other users opinions.
This type of models have shown promising results compared to classic statistical analyzes cited previously. 
We propose a novel graph, feature-based machine learning model to face the cold-start item scenario, 
learning the relevance of features from probabilities of item-based collaborative filtering algorithms.
\end{abstract}


\maketitle

\section{Graph-based models}
Defining $D$ as the dataset of a Recommender System composed by the set of users $U$, the set of items $I$, 
the set of item features $F$ and all the relations among them. Defining $V = U \cup I \cup F$ as the set 
of vertices, $E \subseteq (U \times I) \cup (I \times U) \cup (I \times F) \cup (F \times I)$ as the set 
of edges and assigning to every edge a weight equal to the value of the relation between the nodes it connects,
we can represent $D$ with a weighted tripartite graph $G = (V, E)$.
Recommendations on $D$ can be provided exploiting random walks over $G$, accomplished by starting on a vertex 
and choosing randomly one of its neighbors at each step. 
We can represent $G$ with a square $|V| \times |V|$ adjacency matrix $A$, where every entry 
$A[i,j], i,j \in V$ represents the weight $w(i,j)$ of the edge that connects the node $i$ to $j$.
Normalizing $A$ rowwise, we can compute the probability, or markovian, matrix P where 
$P[i,j]$ represents the probability of choosing any node $j \in V$ when standing in any node $i \in V$.
To get the probability relative to random walks of length $l$, we can elevate $P$ to the power of $l$.
Finally, recommendation lists are provided selecting, for each users, the items with highest probabilities. 
It has been shown in literature that short random walks achieve best performance in most situations\cite{cooper2014}.
In particular, if we consider paths of length 3 over $G$, we can distinguish two types, starting from a user node:
\begin{itemize}
\item \textit{user-item-user-item}: a \textit{collaborative path} that exploits only interactions to reach 
	the destination. Its results are obtained removing edges from items to features before elevating P.
    This approach has been called $P^3$\cite{cooper2014}.
\item \textit{user-item-feature-item}: a \textit{content-based path} that exploits item features over other 
	users' interactions to reach the destination. Its results are obtained removing edges from items to users
    before elevating $P$.
\end{itemize}

\section{Model simplification}
In previous sections we stated that a graph-based random walk model uses only final user-item probabilities 
to produce recommendation lists, which means that of the whole 3 steps random walk result we only 
need the one referred to complete paths from users to items, that is represented by one single 
submatrix of $P$.
This property allows to reduce the exponentiation of $P$ to a multiplication of three of its submatrices. 
We define four non-zero submatrices of $P$ that can be derived directly from the
feedback matrix $URM^{|U| \times |I|}$ and the binary item content matrix $ICM^{|I| \times |F|}$:
\begin{itemize}
\item $P_{ui}$ probability to reach an item node from a user node %in one step
	%\[ P_{ui}[u,i] = \left( \frac{URM[u,i]}{\sum_{j \in I} URM[u,j]} \right) \]

\item $P_{iu}$: probability to reach a user node from an item node %in one step
	%\[ P_{iu}[i,u] = \left( \frac{URM^T[i,u]}{\sum_{v \in U} URM^T[i,v]} \right) \]

\item $P_{if}$: probability to reach a feature node from an item node %in one step
	%\[ P_{if}[i,f] = \left( \frac{ICM[i,f]}{\sum_{f' \in F} ICM[i,f']} \right) \]
    
\item $P_{fi}$: probability to reach an item node from a feature node %in one step
	%\[ P_{fi}[f,i] = \left( \frac{ICM^T[f,i]}{\sum_{j \in I} ICM^T[f,j]} \right) \]   
\end{itemize}
The estimated user-item probabilities used for recommendations, that we will call $URM'$, can be 
obtained with two different multiplications of these submatrices, depending on the nature of the path:
\begin{itemize}
\item \textit{Collaborative}: $ URM' = P_{ui} \odot P_{iu} \odot P_{ui} $
\item \textit{Content-based}: $ URM' = P_{ui} \odot P_{if} \odot P_{fi} $
\end{itemize}

\section{Feature weighting model}
We can introduce feature weights $w_f, f \in F$ over edges that connect items to 
features, obtaining a variant of $P_{if}$ that we will call $P'_{if}$.
This way we can influence the probability to reach feature nodes and, consequently, other item nodes 
in the last steps of content-based paths. 
Note that weights have to be strictly positive, because they directly become probabilities.
As we stated in previous sections, we want to exploit collaborative information to estimate
feature weights, so, in order to do that, we want to obtain as similar results as possible
between the collaborative path and the weighted content-based one. In other words we want that:
\[ P_{ui} \odot P_{iu} \odot P_{ui} = P_{ui} \odot P'_{if} \odot P_{fi} \]
\[ P_{iu} \odot P_{ui} = P'_{if} \odot P_{fi} \]
Now we can define a regression problem over feature weights to solve the equation, minimizing 
the residual sum of squares with Stochastic Gradient Descent.
Given two items $j,k \in I$, we can formalize the problem as:
\[ \argmin_{w} \sum_{j \in I} \sum_{k \in I} ((P'_{if} \odot P_{fi})[j,k] - (P_{iu} \odot P_{ui})[j,k])^2 \]

\section{Target matrix}
We can state that our model is the solution to a regression problem where the target is a $|I| \times |I|$ 
probability matrix obtained with 3-step random walks following the collaborative path.
We will refer to this solution as the hybrid path. 
However, we could use as target any probability matrix that contains collaborative information. 
In particular, we will see the results obtained using the $|I| \times |I|$ probability matrix of the 
$RP^3_\beta$\cite{paudel2017} approach, calculated adapting the popularity-based re-ranking procedure. 
We will refer to this alternative as the \textit{re-ranked} hybrid path.

\section{Datasets}
We tested our model on the well known Movielens 20M dataset, using genres and lemmatized
1 and 2-grams of user based tags as features, and on The Movies Dataset, publicly available on 
Kaggle\footnote{\url{https://www.kaggle.com/rounakbanik/the-movies-dataset}}, that adds to the Full Movielens
dataset the editorial items metadata available on TMDb.
To remove some noise, we applied some filters on both the datasets: we removed items and users with too 
few interactions, items with too few or too many features, and too rare features. 
We split each dataset for train and test, in order to keep a test set that could reproduce
a cold start scenario. So we kept the 20\% of the items, chosen randomly, and all their ratings for the 
test set, while we used the remaining 80\% for the train phase. Then we split the training set in two 
more sets with the same 80-20 proportions, respectively for the training and the validation of the model.

\section{Evaluation}
For the evaluation we used three common metrics of Recommender Systems literature, that can highlight 
the accuracy of the model in prediction and ranking: we used the \textit{@5} variants
of Recall, Mean Average Precision and Normalized Discounted Cumulative Gain.
We compared the results obtained by different approaches:
\begin{itemize}
\item CBF: Content-based KNN algorithm with cosine similarity
\item CBF-IDF: Content-based KNN algorithm with cosine similarity, using IDF to assign feature weights
\item $CP^3$: 3-steps random walks following content-based path
\item $HP^3$: 3-steps random walks following hybrid path
\item $HP^3_R$: 3-steps random walks following hybrid re-ranked path
\end{itemize}

\section{Results}

\begin{table}
	\centering
	\begin{tabular}{ | l | l | l | l | } \hline
      	\thead{Algorithm} & \thead{Recall} & \thead{MAP} & \thead{NDCG} \\ \hline
        CBF & 0.10135 & 0.22026 & 0.13957\\ \hline
        CBF-IDF & 0.10147 & 0.21813 & 0.13941 \\ \hline \hline
        $CP^3$ & 0.08853 & 0.20617 & 0.12359 \\ \hline
        $HP^3$ & 0.09561 & 0.21323 & 0.13247 \\ \hline
        $HP^3_R$ & \textbf{0.10919} & \textbf{0.22826} & \textbf{0.14782} \\ \hline
        %$LFW$ & 0.09849 & 0.19786 & 0.12963 \\ \hline
		%$LFW_R$ & 0.10389 & 0.22567 & 0.14245 \\ \hline
	\end{tabular}
	\caption{Performance @5 of different algorithms on test set of Movielens 20M dataset, the baselines are on top. }
	\label{tab:test5_movielens}

	\begin{tabular}{ | l | l | l | l | } \hline
        \thead{Algorithm} & \thead{Recall} & \thead{MAP} & \thead{NDCG} \\ \hline
        CBF & 0.04850 & 0.08727 & 0.06594 \\ \hline
        CBF-IDF & 0.05033 & 0.08988 & 0.06720 \\ \hline \hline
        $CP^3$ & 0.05482 & 0.09987 & 0.07298 \\ \hline
        $HP^3$ & \textbf{0.06810}  & \textbf{0.12176} & \textbf{0.09118} \\ \hline
        $HP^3_R$ & 0.06776 & 0.11934 & 0.09017 \\ \hline
		%$LFW$ & 0.06715 & 0.10987 & 0.08671 \\ \hline
		%$LFW_R$ & 0.06103 & 0.09786 & 0.07778 \\ \hline
	\end{tabular}
	\caption{Performance @5 of different algorithms on test set of The Movies Dataset, the baselines are on top.}
	\label{tab:test5_themovies}
    \vspace{-8mm}
\end{table}

Analyzing results summarized in Tables \ref{tab:test5_movielens} and \ref{tab:test5_themovies}, we can
see that both the hybrid paths outperform the purely content-based one,
which means that the collaborative information exploited is useful and increases performance.
We can also notice that the target matrix influences the quality of the final model. In particular,
the re-ranked path provides more reliable results, and its performance is higher than both
Content-based KNN approaches on both datasets. The non re-ranked path, instead, is not able to reach
CBF scores on Movielens 20M, but obtains the best results on The Movies Dataset. 
In conclusion, we can state that the model was able to outperform both non-weighted 
and IDF feature weighting approaches, showing the importance of collaborative information and proving 
to be a potentially good solution for cold-start scenarios.

\section{Conclusion}
We proposed a new approach to face the item cold-start problem of Recommender Systems.
We have shown that it is possible to model a hybrid graph-based recommender exploiting collaborative 
information to estimate feature weights and improve quality of content-based recommendations.
Future directions include validating this results on more datasets and baselines, as well as learning from other collaborative probability matrices.

\bibliographystyle{ACM-Reference-Format}
\bibliography{bibliography}

\end{document}